\PassOptionsToPackage{backref=page}{hyperref}
\PassOptionsToPackage{table}{xcolor}
\documentclass{article}

\usepackage[T1]{fontenc}
\usepackage{microtype}
\usepackage{graphicx}
\usepackage{subfigure}
\usepackage{booktabs}
\usepackage{tabularx}
\usepackage{enumitem}
\usepackage{longtable}

\usepackage{hyperref}

\usepackage[accepted]{icml2023} \usepackage{amsmath}
\usepackage{amssymb}
\usepackage{mathtools}
\usepackage{amsthm}
\usepackage{mathtools}  \usepackage{nicefrac}
\usepackage{xspace}
\usepackage{pifont}

\usepackage{pgfplots}
\pgfplotsset{compat=1.16}
\usepgfplotslibrary{fillbetween, groupplots}

\usepackage{todonotes} 

 \theoremstyle{plain}
\newtheorem{theorem}{Theorem}[section]

\theoremstyle{definition}
\newtheorem{definition}[theorem]{Definition}

\theoremstyle{remark}

\DeclareRobustCommand{\bigO}{\ifmmode
         \mathcal{O}
    \else
        \GenericError{}{Attempt to use bigO outside of math mode}\fi
}
\makeatother

\DeclareRobustCommand{\bigpolyO}{\ifmmode
         \Tilde{\mathcal{O}}
    \else
        \GenericError{}{Attempt to use bigO outside of math mode}\fi
}
\makeatother

\renewcommand*{\backrefalt}[4]{\ifcase #1 No citations.\or
Cited on page #2.\else
Cited on pages #2.\fi
}

\newcommand*{\mA}{\mathbf{A}}
\newcommand*{\mC}{\mathbf{C}}

\newcommand*{\mI}{\mathbf{I}}

\newcommand*{\mM}{\mathbf{M}}
\newcommand*{\mLambda}{\mathbf{\Lambda}}

\newcommand*{\mP}{\mathbf{P}}
\newcommand*{\mSigma}{\mathbf{\Sigma}}
\newcommand*{\mU}{\mathbf{U}}
\newcommand*{\mV}{\mathbf{V}}
\newcommand*{\mW}{\mathbf{W}}
\newcommand*{\mX}{\mathbf{X}}
\newcommand*{\mY}{\mathbf{Y}}
\newcommand*{\mZ}{\mathbf{Z}}

\newcommand*{\sR}{\mathbb{R}}

\newcommand*{\sS}{\mathbb{S}}

\newcommand*{\vB}{\mathbf{b}}
\newcommand*{\vX}{\mathbf{x}}
\newcommand*{\vY}{\mathbf{y}}

\newcommand*{\expect}{\mathbb{E}}

\definecolor{cycle1}{RGB}{235,172,35}
\definecolor{cycle2}{RGB}{184,0,88}
\definecolor{cycle3}{RGB}{0,140,249}
\definecolor{cycle4}{RGB}{0,110,0}
\definecolor{cycle5}{RGB}{0,187,173}
\definecolor{cycle6}{RGB}{209,99,230}
\definecolor{cycle7}{RGB}{178,69,2}
\definecolor{cycle8}{RGB}{255,146,135}
\definecolor{cycle9}{RGB}{89,84,214}
\definecolor{cycle10}{RGB}{0,198,248}
\definecolor{cycle11}{RGB}{135,133,0}
\definecolor{cycle12}{RGB}{0,167,108}
\definecolor{cyclegray}{RGB}{189,189,189}

\newtheoremstyle{break}
  {\topsep}{\topsep}{\itshape}{}{\bfseries}{}{\newline}{}\theoremstyle{break}

\newcommand*{\bad}{\textcolor{cycle2}{\ding{55}}}
\newcommand*{\good}{\textcolor{cycle4}{\ding{51}}}
\newcommand*{\ok}{\textcolor{cycle6}{\ding{96}}} 
\begin{document}

\twocolumn[
\icmltitle{Unsupervised Embedding Quality Evaluation}
\icmltitlerunning{Unsupervised Embedding Quality Evaluation}

\icmlsetsymbol{equal}{*}

\begin{icmlauthorlist}
\icmlauthor{Anton Tsitsulin}{goog}
\icmlauthor{Marina Munkhoeva}{tueb}
\icmlauthor{Bryan Perozzi}{goog}
\end{icmlauthorlist}

\icmlaffiliation{goog}{Google Research, New York, USA}
\icmlaffiliation{tueb}{Max Planck Institute for Intelligent Systems, T{\"u}bingen, Germany}

\icmlcorrespondingauthor{Anton Tsitsulin}{tsitsulin@google.com}

\icmlkeywords{Graph Machine Learning, Machine Learning, GNNs}

\vskip 0.3in
]

\printAffiliationsAndNotice{} \begin{abstract}
Unsupervised learning has recently significantly gained in popularity, especially with deep learning-based approaches.
Despite numerous successes and approaching supervised-level performance on a variety of academic benchmarks, it is still hard to train and evaluate SSL models in practice due to the unsupervised nature of the problem.
Even with networks trained in a supervised fashion, it is often unclear whether they will perform well when transferred to another domain.

Past works are generally limited to assessing the amount of information contained in embeddings, which is most relevant for self-supervised learning of deep neural networks.
This works chooses to follow a different approach: can we quantify how easy it is to linearly separate the data in a stable way?
We survey the literature and uncover three methods that could be potentially used for evaluating quality of representations.
We also introduce one novel method based on recent advances in understanding the high-dimensional geometric structure of self-supervised learning.

We conduct extensive experiments and study the properties of these metrics and ones introduced in the previous work.
Our results suggest that while there is no free lunch, there are metrics that can robustly estimate embedding quality in an unsupervised way.
\end{abstract} \section{Introduction}\label{sec:introduction}

With proliferation of unsupervised and self-supervised deep learning methods in the recent years, there is an increasing need to quantify the quality of representations produced by such methods.
Across different domains, this is commonly done with training linear classifiers (\emph{probes}) against known labels~\cite{perozzi2014deepwalk,chen2020simple}.
However, in unsupervised settings \emph{there are no labels} to begin with.
How can we do model selection, optimize methods' hyperparameters, or even verify the method worked at all?

In search of such metrics, we turn our attention to different sub-fields of numerical linear algebra, machine learning and optimization, and high-dimensional probability.
We identify three promising candidate metrics and introduce one based on the expected distribution of embedding distances.
We then proceed to test them on two conceptually novel domains: \emph{supervised} model selection and shallow single-layer graph embedding learning.

Our experimental results indicate there is no ``free lunch''---a metric that is universally dominating---thus calling for a comprehensive suite of evaluation metrics.
Despite that, metrics introduced in this work exhibit, like stable rank and coherence, display stronger correlation to downstream task performance of the supervised models, are more computationally stable, and suit shallow embedding models much better than state-of-the-art ones.

We summarize our key contributions as follows:
\begin{itemize}[noitemsep,topsep=-8pt,parsep=0pt,partopsep=0pt]
    \item We identify three different perspectives on evaluation of embedding quality in unsupervised manner and introduce four metrics based on these perspectives.
    \item We experimentally study two novel settings for embedding quality evaluation, showing that standard metrics often fail when shallow models are being studied.
    \item We conduct a study on computational stability of all metrics and identify the minimum viable sample sizes.
    \item We demonstrate that the proposed metrics are at least as effective as state-of-the-art ones in terms of downstream quality prediction while having more intuitive behavior for shallow embedding models.
\end{itemize}
\vspace{-0mm} \section{Related Work}\label{sec:relatedwork}

The literature on evaluating representations in unsupervised way is still sparse.
Arguably, \emph{dimensional collapse}~\cite{hua2021feature} has sparked initial interest in the area.
In dimensional collapse, some dimensions become non-meaningful (collapse) during training.
Because of that problem, three concurrent metrics, which we introduce below, all study the problem of measuring such collapse from different angles.

\paragraph{$\alpha$-ReQ}\cite{agrawal2022alpha} fits a power-law to the singular values of representations, meaning $\lambda_i\varpropto i^{-\alpha}$.
Logarithmic decay of the spectrum with slope $\alpha=1$ was recently proven to provide the best generalization in infinite-dimensional analysis of linear regression~\cite{bartlett2020benign}.
In practice, a simple linear regression estimator on a log-log scale is used to estimate the value of $\alpha$.
This approach for estimating the power-law exponent is considered inaccurate~\cite{clauset2009power}.

\paragraph{RankMe}\cite{garrido2022rankme,roy2007effective} is a method based on estimating the effective rank of a matrix.
In a strict numerical linear algebraic sense, most embedding matrices are full-rank.
``Softer'' definitions allow to capture not only fully collapsed dimensions but also general underutilization of the parameter space.

\begin{definition}
Given a matrix $\mM \in \sR^{n_1 \times n_2}$ with SVD $\mM = \mU \mSigma \mV^\top$, its
effective rank is the entropy of its normalized singular values, defined as
$$
\textrm{RankMe}(\mM) = - \sum_i p_i\log p_i, \quad p_i = \frac{\sigma_i}{\lVert\mSigma\rVert_1}.
$$
\end{definition}

\paragraph{NESum}\cite{he2022exploring} analyzes eigenspectrum of the covariance matrix of representations.
It is introduced as a heuristic metric complementing the analysis of features learned by the barlow twins loss~\cite{zbontar2021barlow}.

\begin{definition}
Given a matrix $\mM \in \sR^{n_1 \times n_2}$ with covariance that can be decomposed as $\mC = \mU \mLambda \mU^\top$:
$$
\textrm{NESum}(\mM) = \sum_i\frac{\lambda_i}{\lambda_0},
$$
with convention of $\frac{0}{0}=0$.
\end{definition} \section{Three Perspectives on Embedding Quality}\label{sec:methods}

We now study three different perspectives on estimating embedding ``quality''.
All measures we have discussed so far aim to answer an information-theoretic question on representations: \emph{Do embedding carry as much information as their size allows?}
However, there are different questions worth answering.
This paper introduces four novel metrics for embedding quality evaluation based on different perspectives on the embedding quality.

The following section pursues the linear classifier perspective on representation quality~\cite{mohri2011can}.
It asks:
\emph{How hard it is to find a suitable transformation from the representations to the targets of the downstream task?}
We show that this is an inherent property of the representations themselves (and the target matrix too, if it's not a classification task).

\subsection{Linear Classifier Perspective}

Let our downstream task be a classification with a target matrix $\mY \in \{0,1\}^{n \times c}$ and a linear probe $h = \mX\mW+\vB$ with weight matrix $\mW$ and bias vector $\vB$. In what follows, we argue that it is easier to find $h$ that yields high accuracy when applied to the input matrix $\mX$ with higher coherence. 

Without loss of generality, we can drop the bias term. For the ease of exposition, we will adopt the Mean-Squared Error loss  (${\mathcal{L} = || \mY - \mX \mW||^2_F}$) for a downstream task.
The optimal weight matrix will then depend on the target and representation matrices, i.e.\ from the derivative condition ${\mX^\top \mY = \mX^\top\mX \mW}$. Given some ${\mA \in \text{ker}(\mX)}$, i.e. a matrix comprised of vectors from the null space of $\mX$, we rewrite the condition as ${\mX^\top \mY = \mX^\top (\mA + \mX^\dagger\mY)}$ and get ${\mW^* = \mX^\dagger \mY + \mA}$ for any ${\mA \in \text{ker}(\mX)}$.

Assuming we can always find an optimal weight matrix, to minimize the loss $\mathcal{L}$, the representations $\mX$ should be aligned with the target matrix $\mY$, i.e. the left singular vectors $\mU$ of $\mX = \mU \mSigma \mV$ should span $\mU_\mY$ of $\mY = \mU_\mY \mSigma_\mY \mV_\mY$, where $\mV_\mY = \mI_c$ when $\mY$ is a classification target matrix.

Plugging in the optimal $\mW^*$ into the loss,
\begin{align*}
    || \mY - \mX (\mX^\dagger \mY + \mA) ||^2_F 
    &=  \| \mY - \mU \mSigma \mSigma^\dagger \mU^\top \mY \|^2_F \\
    &= \| (\mI - \mU \mI_d \mU^\top) \mY \|_F^2 \\
    &= \| (\mI - \mI_d)\mU^\top \mY \|_F^2 \\
    &= \| \mY\|_F^2 - \|\mU_d^\top \mU_\mY \mSigma_\mY\|_F^2,
\end{align*}
where ${\mI_d \in \sR^{n\times n}}$ with $d$ ones on the diagonal, and the minimum is reached whenever columns in $\mU$ are aligned with columns in $\mU_\mY$.

Intuitively, if the representation dimensionality is larger than number of classes in the downstream task, i.e. $d > c$, and $\mX$ has full rank (a consequence of most methods being spectral embedding), then the representation basis covers the target basis with high probability.
However, to quantify the extent of this coverage, we will need 
to introduce a notion of incoherence.

\begin{definition}[$\mu_0$-incoherence]
Given matrix $\mM \in \sR^{n_1 \times n_2}$ with rank-$r$ and SVD $\mM = \mU \mSigma \mV^\top$,
$\mM$ is said to satisfy the \emph{standard incoherence} condition with parameter $\mu_0$ if 
$$
    \max_{1 \leq i \leq n_1} ||\mU^\top e_i||_2 \leq \sqrt{\frac{\mu_0 r}{n_1}}, ~
    \max_{1 \leq i \leq n_2} ||\mV^\top e_j||_2 \leq \sqrt{\frac{\mu_0 r}{n_2}},
$$

where $e_i$ is the $i$-th standard basis vector of a respective dimension. Note that $1 \leq \mu_0 \leq \nicefrac{\max(n_1, n_2)}{r}$.
\end{definition}

Informally, standard incoherence characterizes the extent of alignment of the singular vectors to the standard basis.
Incoherence is typically used in low-rank matrix completion problems to estimate a complexity of matrix recovery~\cite{mohri2011can}. In our setting, \emph{lower} incoherence will be indicative of high alignment with target matrix and, thus, \emph{better} performance.

Ideally, if we had access to the targets, we could use joint incoherence $\mu_1(\mZ, \mY)$ to measure the alignment directly. More practical is the case when true labels are not available. There, we will need to rely on the standard coherence $\mu_0(\mZ)$ which measures alignment to the standard basis. Our experiments show that there is indeed a correlation between standard incoherence of the representations and performance on the downstream tasks (almost perfect in some cases).

\subsection{Numerical Linear Algebra Perspective}

Numerical linear algebra provides us with more tools for analysing behaviors of linear classifiers.
One of the classic ones is the condition number, or, in the case of non-square matrices, its generalized version~\cite{ben1966error}.
For example, $\kappa_2$ is used to detect multicollinearity in linear and logistic regression~\cite{belsley2005regression}.

\begin{definition}
Pseudo-condition number of a matrix $\mM$ with SVD $\mM = \mU\mSigma\mV^\top$ is defined as
$$
\kappa_p(\mM) = \lVert \mM \rVert_p \lVert \mM^\dagger \rVert_p \stackrel{p=2}{=} \frac{\sigma_1}{\sigma_n}.
$$
\end{definition}

We are particularly interested in $\kappa_2$, since it is easily computable with SVD, as the pseudo-inverse of $\mM$ is $(\mM^\top\mM)^{-1}\mM=\mU\mSigma^{-1}\mV^\top$, meaning $\lVert \mM^\dagger \rVert_2=\nicefrac{1}{\sigma_n}$.

In the analysis of linear regression, $\kappa_2$ can be used to bound the sensitivity of the system to the change in the input.
Consider a linear system $(\mA+\Delta\mA)\hat{\vX} = \vB$ and its perturbed version $\mA\hat{\vX} = \vB+\Delta\vB$.
Then,
$$
\frac{\lVert\hat{\vX} - \vX\rVert}{\lVert\vX\rVert} \leq 
\frac{\kappa(\mA)}{1-\kappa(\mA)\frac{\lVert\Delta\mA\rVert}{\lVert\mA\rVert}}\left(\frac{\lVert\Delta\mA\rVert}{\lVert\mA\rVert}+\frac{\lVert\Delta\vB\rVert}{\lVert\vB\rVert}\right).
$$

We use $\kappa_2$ to measure stability of learned representations.

\subsubsection{Stable Rank}

Stable rank (also called \emph{effective} rank or intrinsic dimension of a matrix) is another fundamental quality in numerical analysis of random matrices.

\begin{definition}Numerical rank of a matrix $\mM$ is defined as
$$
r(\mM) = \frac{\lVert\mM\rVert_F}{\lVert\mM\rVert^2_2}
$$
\end{definition}

Note that $r(\mM)\leq \textrm{rank}(\mM)$, and that bound is sharp.
Stable rank is a useful tool that guides fundamental numerical problems, including matrix sampling and covariance estimation.
Let us restate Theorem 1.1 from~\citet{rudelson2007sampling}:

\begin{theorem}
Let $\mA$ be an $n \times d$ matrix with stable rank $r$. Let $\varepsilon, \delta \in (0, 1)$, and let $m \leq n$ be an integer such that
$$
m \geq C \left(\frac{r}{\varepsilon^4\delta}\right) \log\left(\frac{r}{\varepsilon^4\delta}\right).
$$
Consider a $m \times d$ matrix $\tilde{\mA}$, which consists of $m$ normalized rows of $\mA$ picked independently with replacement, with probabilities proportional to the squares of their Euclidean lengths. Then with probability at least $1 - 2\mathrm{exp}(-\nicefrac{c}{\delta})$ the following holds. For a positive integer $k$, let $\mP_k$ be the orthogonal projection onto the top k left singular vectors
of $\tilde{\mA}$. Then,
$$
\lVert\mA - \mA\mP_k\rVert = \sigma_{k+1}(\mA) + \varepsilon\lVert\mA\rVert_2.
$$
\end{theorem}

This suggests that the numerical rank determines how hard it is to estimate the matrix by subsampling its rows.
Intuitively, a well-distributed representations should be hard to estimate; we will observe that this is indeed the case in practice.

\subsection{High-dimensional Probability Perspective}

In self-supervised learning, \citet{assran2023hidden} shows that several contrastive learning methods try to distribute representations equally in the space.
High-dimensional probability can provide us with an estimate of pairwise distances when embeddings are distributed uniformly on a $d$-dimensional unit sphere $\sS^d$.

Given $L_2$ normalized embeddings $\mW \in \sR^{n \times d}$, a measure of clustering can be defined using the norm of the pairwise dot product matrix $Q = \lVert\mW\mW^\top\rVert_F$. Since the expected dot product of high-dimensional isotropic random vectors $\langle \vX, \vY \rangle \asymp \frac {1}{n}$~\citep[Remark 3.2.5]{vershynin2018high}, we can estimate $\expect[Q] = n + \frac{n(n-1)}{d} $.
The maximum metric value $Q=n^2$ can only be achieved in the collapsed case.
Combining all normalizations to get a metric upper-bounded that is upper-bounded by 1, we get:
\begin{definition}
\begin{align*}
    \textrm{SelfCluster}(\mW) &= \frac{\lVert \mW\mW^\top\rVert_F-n-\frac{n*(n-1)}{d}}{n^2 - n - \frac{n*(n-1)}{d}} \\
    &= \frac{d\lVert \mW\mW^\top\rVert_F - n (d+n-1)}{(d-1)(n-1)n}.
\end{align*}
\end{definition}

SelfCluster allows us to estimate how much the embeddings are clustered in the embedding space compared to random distribution on a sphere.
The downside of this metric is the requirement of pairwise computations, which is expensive for large number of points.
We now proceed to study the proposed metrics on real-world data.

 \section{Experiments}\label{sec:experiments}

In contrast to previous work~\cite{agrawal2022alpha,garrido2022rankme}, we shift our attention from self-supervised learning to novel, more generally applicable settings.
We experimentally study proposed metrics on two novel use-cases: (i) supervised representation learning with deep neural networks and (ii) unsupervised graph embeddings.
Supervised representation learning allows us to gain insights into performance of semi-supervised learning systems.
Graph embedding, on the other hand, has very different architecture---shallow single-layer network---and optimization.

Section~\ref{ssec:exp:supervised:stability} further provides a novel study on computational stability of different embedding quality evaluation metrics.
Stability is important for many practical application, since the most computationally stable metrics can be even computed during training for monitoring purposes.

\subsection{Supervised Network Performance Prediction}\label{ssec:exp:supervised}

We used~\citet{rw2019timm} repository of supervised PyTorch models, accessed May 2023.
\cite{deng2009imagenet}
We ran inference of all available models, as permitted by GPU memory, on the validation set, and a subset of models\footnote{Full list available in the Appendix.}---on the full training set.
Inference was performed on a single 16-core machine with NVIDIA RTX 4090 and 64Gb RAM.

\begin{figure}[t]
\centering
\begin{tikzpicture}
\begin{groupplot}[group style={
                      group name=myplot,
                      group size= 2 by 4, horizontal sep=0.75cm,vertical sep=1.15cm},height=3.7cm,width=0.55\linewidth,title style={at={(0.5,0.9)},anchor=south},every axis x label/.style={at={(axis description cs:0.5,-0.15)},anchor=north},]
\nextgroupplot[
 	title = {{$\alpha$-ReQ}, $\rho=0.40$},
	ylabel=ACC$\times100$,
	xmode=log,
]
\addplot[only marks, color=cycle3,fill opacity=0.4,draw opacity=1] table[x=alpha_req,y=acc] {figures/data/imagenet-train.tex};

\nextgroupplot[
 	title = {{NESum}, $\rho=0.40$},
]
\addplot[only marks, color=cycle3,fill opacity=0.4,draw opacity=1] table[x=ne_sum,y=acc] {figures/data/imagenet-train.tex};

\nextgroupplot[
 	title = {{RankMe}, $\rho=0.21$},
	ylabel=ACC$\times100$,
]
\addplot[only marks, color=cycle3,fill opacity=0.4,draw opacity=1] table[x=rankme,y=acc] {figures/data/imagenet-train.tex};

\nextgroupplot[
 	title = {{RankMe$^*$}, $\rho=0.00$},
]
\addplot[only marks, color=cycle3,fill opacity=0.4,draw opacity=1] table[x=rankme_adj,y=acc] {figures/data/imagenet-train.tex};

\nextgroupplot[
 	title = {\textbf{\textcolor{cycle2!50!black}{Stable rank}}, $\rho=0.12$},
	ylabel=ACC$\times100$,
]
\addplot[only marks, color=cycle3,fill opacity=0.4,draw opacity=1] table[x=numerical_rank,y=acc] {figures/data/imagenet-train.tex};

\nextgroupplot[
 	title = {\textbf{\textcolor{cycle2!50!black}{Cond.\ Num.}}, $\rho=-0.48$},
 	xmode=log,
]
\addplot[only marks, color=cycle3,fill opacity=0.4,draw opacity=1] table[x=pseudo_condition_number,y=acc] {figures/data/imagenet-train.tex};

\nextgroupplot[
 	title = {\textbf{\textcolor{cycle2!50!black}{Coherence}}, $\rho=0.30$},
 	xmode=log,
	ylabel=ACC$\times100$,
]
\addplot[only marks, color=cycle3,fill opacity=0.4,draw opacity=1] table[x=coherence,y=acc] {figures/data/imagenet-train.tex};
\end{groupplot}
\end{tikzpicture}
\vspace{-5mm}
\caption{Representation quality metrics on the ImageNet training set for over 30 pre-trained models. Spearman rank correlation $\rho$ to the test set accuracy displayed per metric in the title. Methods introduced in this work are highlighted in \textbf{\textcolor{cycle2!50!black}{colored bold}}.\label{fig:imagenet-train}}
\end{figure}
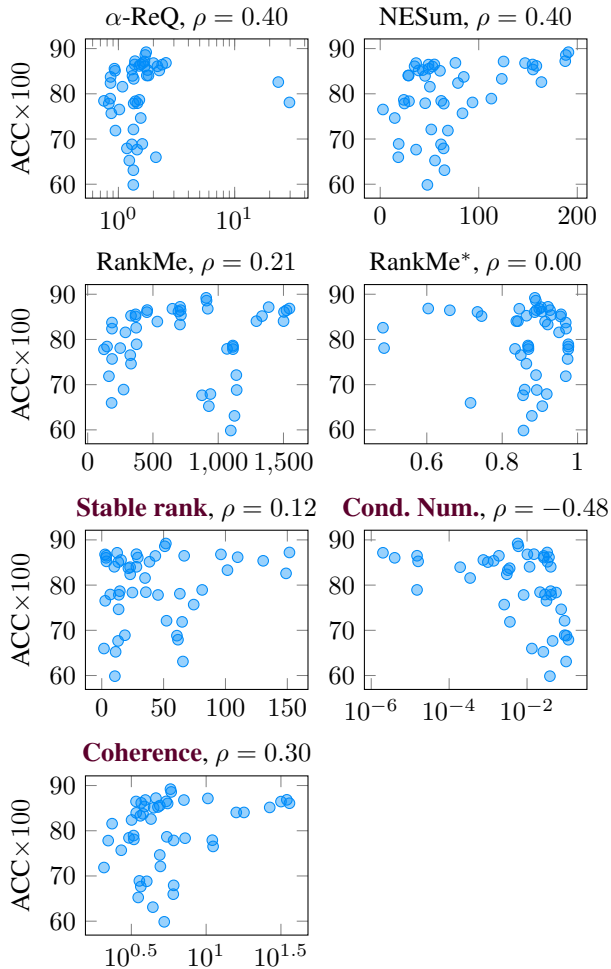 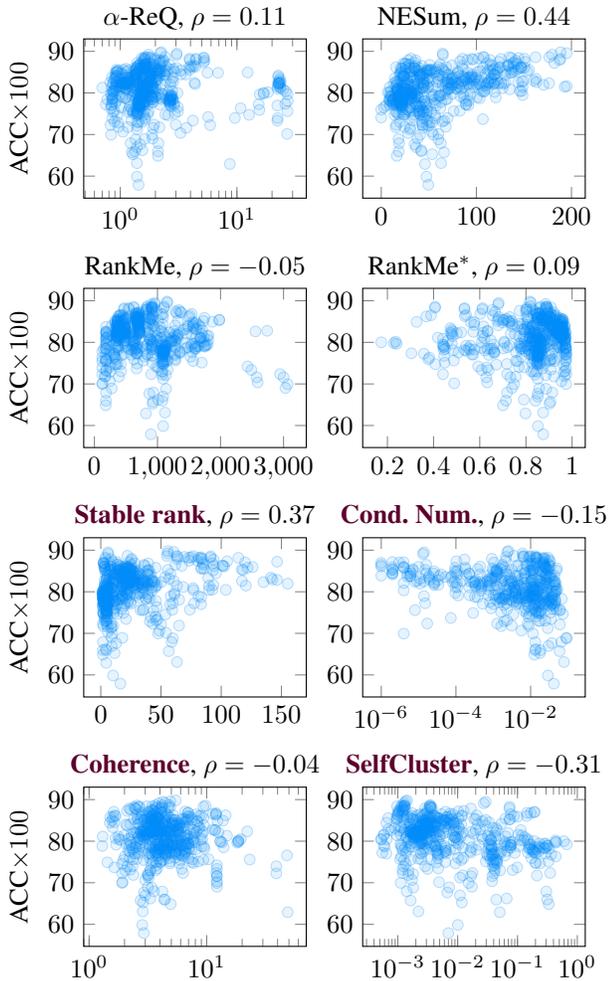
\begin{figure}[t]
\centering
\begin{tikzpicture}
\begin{groupplot}[group style={
                      group name=myplot,
                      group size= 2 by 4, horizontal sep=0.75cm,vertical sep=1.2cm},height=3.7cm,width=0.55\linewidth,title style={at={(0.5,0.9)},anchor=south},every axis x label/.style={at={(axis description cs:0.5,-0.15)},anchor=north},]
\nextgroupplot[
 	title = {{$\alpha$-ReQ}, $\rho=0.11$},
	ylabel=ACC$\times100$,
	xmode=log,
]
\addplot[only marks, color=cycle3,fill opacity=0.1,draw opacity=0.3] table[x=alpha_req,y=acc] {figures/data/imagenet-valid.tex};

\nextgroupplot[
 	title = {{NESum}, $\rho=0.44$},
]
\addplot[only marks, color=cycle3,fill opacity=0.1,draw opacity=0.3] table[x=ne_sum,y=acc] {figures/data/imagenet-valid.tex};

\nextgroupplot[
 	title = {{RankMe}, $\rho=-0.05$},
	ylabel=ACC$\times100$,
]
\addplot[only marks, color=cycle3,fill opacity=0.1,draw opacity=0.3] table[x=rankme,y=acc] {figures/data/imagenet-valid.tex};

\nextgroupplot[
 	title = {{RankMe$^*$}, $\rho=0.09$},
]
\addplot[only marks, color=cycle3,fill opacity=0.1,draw opacity=0.3] table[x=rankme_adj,y=acc] {figures/data/imagenet-valid.tex};

\nextgroupplot[
 	title = {\textbf{\textcolor{cycle2!50!black}{Stable rank}}, $\rho=0.37$},
	ylabel=ACC$\times100$,
]
\addplot[only marks, color=cycle3,fill opacity=0.1,draw opacity=0.3] table[x=numerical_rank,y=acc] {figures/data/imagenet-valid.tex};

\nextgroupplot[
 	title = {\textbf{\textcolor{cycle2!50!black}{Cond.\ Num.}}, $\rho=-0.15$},
 	xmode=log,
]
\addplot[only marks, color=cycle3,fill opacity=0.1,draw opacity=0.3] table[x=pseudo_condition_number,y=acc] {figures/data/imagenet-valid.tex};

\nextgroupplot[
 	title = {\textbf{\textcolor{cycle2!50!black}{Coherence}}, $\rho=-0.04$},
 	xmode=log,
	ylabel=ACC$\times100$,
]
\addplot[only marks, color=cycle3,fill opacity=0.1,draw opacity=0.3] table[x=coherence,y=acc] {figures/data/imagenet-valid.tex};
\nextgroupplot[
 	title = {\textbf{\textcolor{cycle2!50!black}{SelfCluster}}, $\rho=-0.31$},
 	xmode=log,
]
\addplot[only marks, color=cycle3,fill opacity=0.1,draw opacity=0.3] table[x=self_clustering,y=acc] {figures/data/imagenet-valid.tex};

\end{groupplot}
\end{tikzpicture}
\vspace*{-5mm}
\caption{Representation quality metrics on the ImageNet validation set of over 1000 pre-trained models. Spearman rank correlation $\rho$ to the test set accuracy displayed per metric in the title. Methods introduced in this work are highlighted in \textbf{\textcolor{cycle2!50!black}{colored bold}}.\label{fig:imagenet-val}}
\vspace*{-3mm}
\end{figure} 
\subsubsection{Downstream Quality Correlation}\label{sssec:exp:supervised:quality}

Figures~\ref{fig:imagenet-train} and~\ref{fig:imagenet-val} present rank correlation of the different embedding quality metrics to downstream prediction quality on ImageNet, measured for training and validation set embeddings respectively.
We do not report SelfCluster metric results on the training set because of its quadratic time complexity.
Since RankMe is dependent on the dimensionality of the data, we normalize its values and call the metric RankMe$^*$.
This new metric has the range between 0 and 1, and represents relative utilization of the embedding space.

On the training set evaluation, $\alpha$-ReQ, NESum, pseudo-condition number, and coherence all show significant correlation to the test set performance.
Out of these metrics, $\alpha$-ReQ is the only metric with significant outliers, possibly due to the power law estimation issues~\cite{clauset2009power}.
High stable rank, NESum, and coherence seem to indicate good test test performance of the model.
Note that the models we selected for training set evaluation are pareto-optimal in terms of either parameter size or inference speed.
This allowed us to significantly restrict the model set size without affecting representativeness of selected models.

On the validation set performance with expanded model set, the correlation between many metrics and test set performance drops to near-zero.
This can be attributed to both expanded model set, which has many under-performing models as well as the general instability of the computation on the smaller example set.
We further examine the computational stability considerations in the next section.
Only NESum, stable rank and self clustering achieve significant correlation to the test set performance.
Across both training and validation sets, NESum demonstrates strong downstream performance correlation while both variants of RankMe are not able to successfully predict supervised task performance.

\begin{table}[tb]
\centering
\setlength{\aboverulesep}{0pt}
\newcolumntype{C}{>{\raggedleft\arraybackslash}X}
\newcolumntype{S}{>{\centering\arraybackslash\hsize=.5\hsize}X}
\caption{\label{tbl:stability}Batch sizes needed to achieve constant multiplicative approximation factors compared to evaluation on the full ImageNet training set on XX networks. Additionally, we check that each metric lower-bounds the true value. The result can be either \good yes, \bad no, or \ok 0.95-approximately. }
\begin{tabularx}{\linewidth}{p{1.5cm}>{\centering\arraybackslash}p{1.3cm}CCCCC}
\toprule
\multicolumn{2}{c}{} & \multicolumn{4}{c}{\emph{Approximation factor}} \\
\cmidrule{3-6}
\emph{metric} & \mbox{Bounded} & 0.5 & 0.7 & 0.9 & 0.95\\
\midrule
$\alpha$-ReQ & \bad & 512 & 4096 & 32768 & --- \\
NESum & \ok & 1024 & 2048 & 8192 & 32768 \\
RankMe & \good & 2048 & 2048 & 8192 & 16384 \\
\midrule
\mbox{Stable rank} & \ok & 512 & 2048 & 8192 & 16384 \\
\mbox{Cond.\ number} & \bad & 4096 & 4096 & 32768 & 65536 \\
Coherence & \good & --- & --- & --- & --- \\
\bottomrule
\end{tabularx}
\vspace*{-2mm}
\caption{\label{tbl:datasets}Dataset statistics. We report total number of nodes $|V|$, average node degree $\Bar{d}$, number of labels $|Y|$.}
\begin{tabularx}{\linewidth}{@{}p{1.85cm}CCCC@{}}
\toprule
\emph{dataset} & $|V|$ & $\Bar{d}$ & $|Y|$ \\
\midrule
Cora & 19793 & 3.20 & 7 \\
Citeseer & 3327 & 1.37 & 6 \\
PubMed & 19717 & 2.25 & 3 \\
\mbox{Amazon PC} & 13752 & 17.88 & 10 \\
\mbox{Amazon Photo} & 7650 & 15.57 & 8 \\
\mbox{MSA-Physics} & 34493 & 7.19 & 5 \\
\mbox{OGB-arXiv} & 169343 & 6.84 & 40 \\
CIFAR-10 & 50000 & 99 & 10 \\
MNIST & 60000 & 99 & 10 \\
\bottomrule
\end{tabularx}
\vspace{-5mm}
\end{table}

\subsubsection{Metric Stability}\label{ssec:exp:supervised:stability}

It is important to have stable metrics for embedding quality evaluation, especially in low-data regimes.
Moreover, if a metric is stable up to very small batch sizes, it can be evaluated during training, greatly enhancing its usability.

To do that, we sample embeddings for ImageNet training set with batch sizes from 128 to 65536, log-space ($2^7$--$2^{16}$) and compare the sampled metric value to the value computed on the whole dataset.
The results are presented in Table~\ref{tbl:stability}.
Numerical rank-based methods are among the most stable, followed by NESum.
One advantage of RankMe over its numerical rank estimation counterpart is that it offers a strong lower-bound in terms of the sample size.
Coherence appears to be strongly data-dependent and least stable.

\subsection{Graph Embedding Quality Prediction}\label{ssec:exp:graph-embedding}

Graph embedding is a common way to solve many tasks arising in the graph mining domain from node classification, link prediction, and community detection.
In the graph embedding process, each node in a graph is mapped to a vector in $\sR^d$, and distances in the embedding space should resemble some similarity metric defined between the nodes in the original graph~\cite{tsitsulin2018verse}.
For an in-depth review of modern graph embedding approaches, readers are referred to~\citet{chami2022machine} survey.

\begin{table*}[!t]
\centering
\setlength{\aboverulesep}{0pt}
\setlength{\belowrulesep}{0pt}
\newcolumntype{C}{>{\raggedleft\arraybackslash}X}
\newcolumntype{S}{>{\centering\arraybackslash\hsize=.5\hsize}X}
\caption{\label{tbl:embedding-perds}Average Spearman rank correlation on two dataset corruption types: na\"ive (N) and component-preserving (C). We highlight datasets where there is a consistent correlation pattern, meaning the same sign and approximately the same magnitude of correlation. Methods proposed in this work exhibit stronger and more consistent correlation patterns across all datasets.}
\begin{tabularx}{\linewidth}{p{1.85cm}CCCCCCCCCC}
\toprule
& \multicolumn{2}{c}{\textbf{Cora}} & \multicolumn{2}{c}{\textbf{Citeseer}} & \multicolumn{2}{c}{\textbf{Pubmed}} & \multicolumn{2}{c}{\textbf{Amazon PC}} & \multicolumn{2}{c}{\textbf{Amazon Photo}} \\
\emph{metric} & \multicolumn{1}{c}{N} & \multicolumn{1}{c}{C} & \multicolumn{1}{c}{N} & \multicolumn{1}{c}{C} & \multicolumn{1}{c}{N} & \multicolumn{1}{c}{C} & \multicolumn{1}{c}{N} & \multicolumn{1}{c}{C} & \multicolumn{1}{c}{N} & \multicolumn{1}{c}{C} \\
\midrule
$\alpha$-ReQ & \cellcolor{cycle4!15}-1.00 &\cellcolor{cycle4!15} -1.00 & \cellcolor{cycle4!15}-1.00 & \cellcolor{cycle4!15}-1.00 & -1.00 & 0.43 & 0.01 & 0.98 & 0.01 & 0.97 \\
NESum & 1.00 & 0.03 & 1.00 & 0.10 & 0.94 & -0.66 & 0.09 & -1.00 & -0.15 & -1.00 \\
RankMe & \cellcolor{cycle4!15}1.00 & \cellcolor{cycle4!15}1.00 & \cellcolor{cycle4!15}1.00 & \cellcolor{cycle4!15}1.00 & 1.00 & -0.37 & -0.05 & -0.99 &\cellcolor{cycle4!15} -0.43 & \cellcolor{cycle4!15}-0.99 \\
\midrule
Stable rank & \cellcolor{cycle4!15}1.00 & \cellcolor{cycle4!15}0.66 & \cellcolor{cycle4!15}1.00 & \cellcolor{cycle4!15}0.30 & \cellcolor{cycle4!15}1.00 & \cellcolor{cycle4!15}0.66 & 0.31 & -1.00 & 0.09 & -1.00 \\
\mbox{Cond.\ number} & \cellcolor{cycle4!15}1.00 & \cellcolor{cycle4!15}0.83 & \cellcolor{cycle4!15}1.00 & \cellcolor{cycle4!15}1.00 & \cellcolor{cycle4!15}1.00 & \cellcolor{cycle4!15}0.26 & 0.20 & -0.99 & 0.10 & -1.00 \\
SelfCluster & \cellcolor{cycle4!15}-1.00 & \cellcolor{cycle4!15}-1.00 & \cellcolor{cycle4!15}-1.00 & \cellcolor{cycle4!15}-0.60 & \cellcolor{cycle4!15}1.00 & \cellcolor{cycle4!15}1.00 & \cellcolor{cycle4!15}1.00 & \cellcolor{cycle4!15}0.99 & \cellcolor{cycle4!15}1.00 & \cellcolor{cycle4!15}1.00 \\ 
Coherence & \cellcolor{cycle4!15}1.00 & \cellcolor{cycle4!15}1.00 & \cellcolor{cycle4!15}0.90 & \cellcolor{cycle4!15}1.00 & \cellcolor{cycle4!15}0.94 & \cellcolor{cycle4!15}1.00 & \cellcolor{cycle4!15}0.99 & \cellcolor{cycle4!15}0.98 & \cellcolor{cycle4!15}0.99 & \cellcolor{cycle4!15}0.98 \\
\midrule
\midrule
\end{tabularx}
\begin{tabularx}{\linewidth}{p{1.85cm}CCCCCCCC}& \multicolumn{2}{c}{\textbf{MSA-Physics}} & \multicolumn{2}{c}{\textbf{OGB-arXiv}} & \multicolumn{2}{c}{\textbf{MNIST}} & \multicolumn{2}{c}{\textbf{CIFAR-10}} \\
\emph{metric} & \multicolumn{1}{c}{N} & \multicolumn{1}{c}{C} & \multicolumn{1}{c}{N} & \multicolumn{1}{c}{C} & \multicolumn{1}{c}{N} & \multicolumn{1}{c}{C} & \multicolumn{1}{c}{N} & \multicolumn{1}{c}{C}  \\
\midrule
$\alpha$-ReQ & -0.70 & 0.94 & -0.81 & 1.00 & -1.00 & 0.98 & \cellcolor{cycle4!15}0.96 & \cellcolor{cycle4!15}0.99 \\
NESum & 0.51 & -0.98 & 0.84 & -1.00 & 0.99 & -0.92 & \cellcolor{cycle4!15}-0.84 &\cellcolor{cycle4!15} -0.99 \\
RankMe & 0.59 & -0.92 & 0.85 & -1.00 & 1.00 & -0.96 & \cellcolor{cycle4!15}-0.94 & \cellcolor{cycle4!15}-1.00 \\
\midrule
Stable rank & 0.52 & -0.97 & 0.99 & -0.99 & 1.00 & -0.78 &\cellcolor{cycle4!15} -0.85 & \cellcolor{cycle4!15}-0.99 \\
\mbox{Cond.\ number} & 0.52 & -0.97 & 0.92 & -1.00 & 1.00 & -0.96 &\cellcolor{cycle4!15} -0.95 &\cellcolor{cycle4!15} -0.99 \\
SelfCluster &\cellcolor{cycle4!15} 0.96 & \cellcolor{cycle4!15}0.98 & \cellcolor{cycle4!15}1.00 &\cellcolor{cycle4!15} 1.00 & \cellcolor{cycle4!15}1.00 & \cellcolor{cycle4!15}1.00 & \cellcolor{cycle4!15}1.00 & \cellcolor{cycle4!15}0.99 \\ 
Coherence & \cellcolor{cycle4!15}0.97 & \cellcolor{cycle4!15}0.99 & \cellcolor{cycle4!15}0.90 & \cellcolor{cycle4!15}1.00 & \cellcolor{cycle4!15}0.89 & \cellcolor{cycle4!15}1.00 & \cellcolor{cycle4!15}0.98 & \cellcolor{cycle4!15}0.99 \\
\bottomrule
\end{tabularx}
\vspace{-3mm}
\end{table*} 
For our experiments, we study representations of the DeepWalk~\cite{perozzi2014deepwalk} model as it is a de-facto standard in the field of unsupervised embedding of graphs with no features.
We use 10 different graph datasets that include both natural and constructed graphs.
We report the dataset statistics in Table~\ref{tbl:datasets} and provide a brief description below:
\begin{itemize}[noitemsep,topsep=0pt,parsep=0pt,partopsep=0pt]
    \item Cora, Citeseer, and Pubmed~\cite{sen2008} are citation networks; nodes represent papers connected by citation edges; features are bag-of-word abstracts, and labels represent paper topics. We use a re-processed version of Cora from~\cite{shchur2018pitfalls} due to errors in the processing of the original dataset.
    \item Amazon~\{PC, Photo\}~\cite{shchur2018pitfalls} are two subsets of the Amazon co-purchase graph for the computers and photo sections of the website, where nodes represent goods with edges between ones frequently purchased together; node features are bag-of-word reviews, and class labels are product category.
    \item OGB-ArXiv~\cite{hu2020open} is a paper co-citation dataset based on arXiv papers indexed by the Microsoft Academic graph. Nodes are papers; edges are citations, and class labels indicate the main category of the paper.
    \item CIFAR and MNIST~\cite{krizhevsky2009learning,lecun1998mnist} are $\varepsilon$-nearest neighbor graphs with $\varepsilon$ such that the average node degree is 100.
\end{itemize}

Instead of changing the parameters of the model, we controllably change the quality of data itself.
We sparsify each graph in two different ways:
\begin{itemize}[noitemsep,topsep=0pt,parsep=0pt,partopsep=0pt]
    \item \textbf{Na\"ive sparsification}: we randomly pick $n\bar{d}$ edges from the original edge set. This method may produce disconnected components, which are known to be difficult to embed correctly.
    \item \textbf{Component-preserving sparsification}: we first ensure the resulting graph is connected by sampling a random spanning tree. Then, we sample $n(\bar{d}-1)$ edges randomly and output the combined graph.
\end{itemize}

It is easy to see both versions create a controllably worse version of the data.
As such, one could expect that representation quality degrades with the sparsity of the input graph, perhaps faster for the na\"ive algorithm, since it does not preserve the component information.
As we will observe later, surprisingly, this is very much not the case for many embedding quality metrics we study.

We sparsify to a fixed number of edges corresponding to a target average node degree from the range $[1.1, 10]$.
Some graphs in our studies have an average node degree $<10$ naturally (cf.\ Table~\ref{tbl:datasets}), in this case, we stop at that number.
We embed each graph 10 times, run a downstream node classification 100 times, and average the result.
We report Spearman rank correlation coefficient $\rho$~\cite{spearman1904proof} between the classification accuracy and each quality metric.

\begin{table}[tb]
\centering
\setlength{\aboverulesep}{0pt}
\newcolumntype{C}{>{\raggedleft\arraybackslash}X}
\newcolumntype{S}{>{\centering\arraybackslash\hsize=.5\hsize}X}
\caption{\label{tbl:embedding-aggregated}Average Spearman rank correlation on two dataset corruption types: na\"ive and component-preserving. We highlight rows where there is a consistent correlation pattern. Two methods introduced in this work strongly and consistently correlate with the downstream classification performance.}
\begin{tabularx}{\linewidth}{p{1.65cm}CC}
\toprule
\emph{metric} & Na\"ive & Connected \\
\midrule
$\alpha$-ReQ & -0.50 & 0.48 \\
NESum & 0.49 & -0.71 \\
RankMe & 0.45 & -0.47 \\
\midrule
Stable rank & 0.56 & -0.46 \\
\mbox{Cond.\ number} & 0.53 & -0.43 \\
\rowcolor{cycle4!15} SelfCluster & 0.55 & 0.60 \\
\rowcolor{cycle4!15} Coherence & 0.95 & 0.99 \\
\bottomrule
\end{tabularx}
\end{table} 
First, we report aggregated results across all datasets in Table~\ref{tbl:embedding-aggregated}.
Surprisingly, most metrics completely revert the correlation sign between two sparsification strategies.
Only SelfCluster and Coherence are aligned with the downstream evaluation, and between them, Coherence displays a near-perfect correlation with the downstream task performance.

\begin{figure*}[!t]
    \centering
    \includegraphics[width=0.84\textwidth]{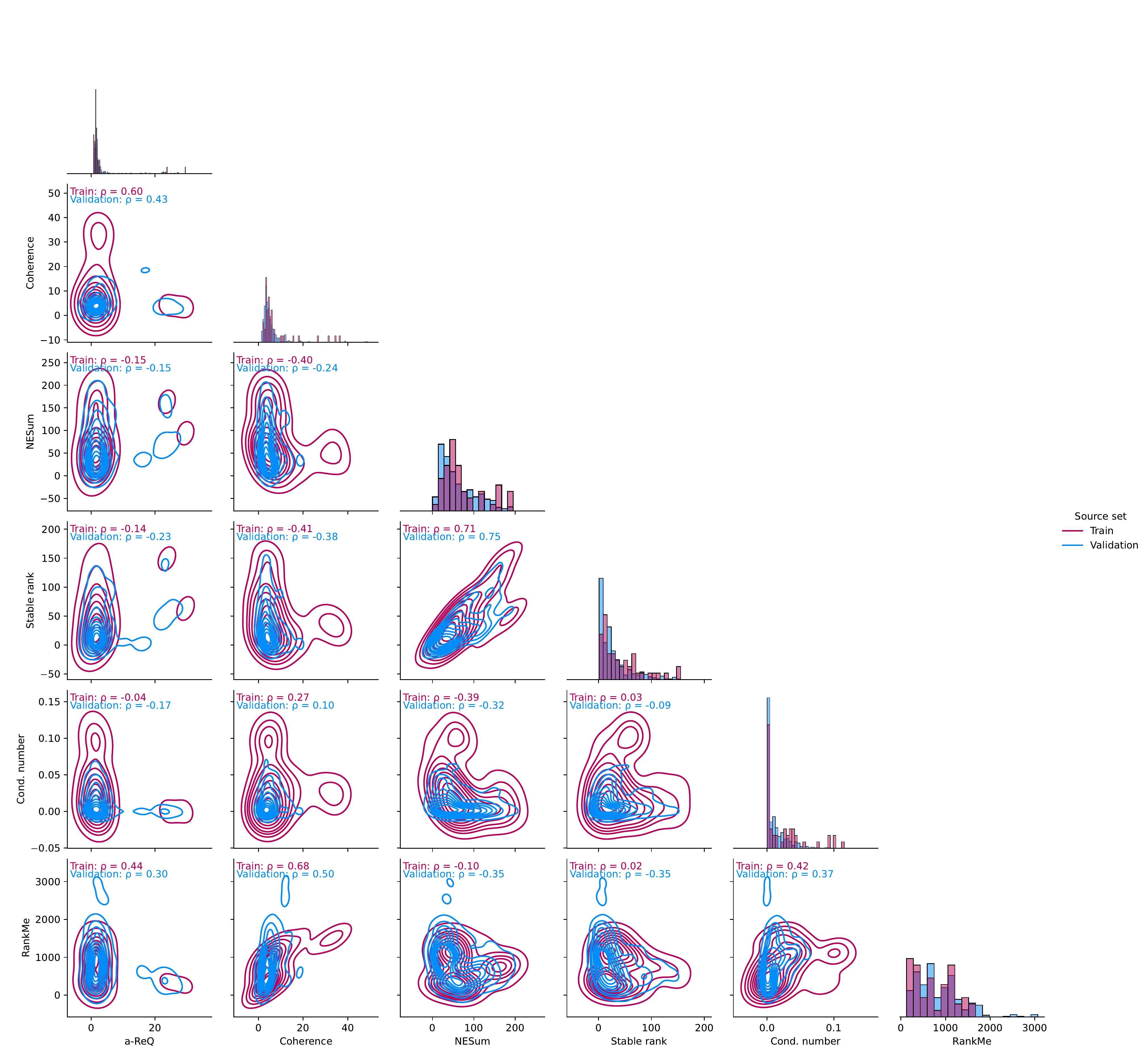}
    \caption{Pairwise density plots of ImageNet representations, as measured on training and validation sets. NEsum is well-correlated to Stable rank. Coherence is moderately correlated to $\alpha$-ReQ and RankMe.  \label{fig:imagenet-pairplot}}
\end{figure*}

Table~\ref{tbl:embedding-perds} provides a more nuanced per-dataset view.
We can observe that while some metrics have strong and consistent correlation patterns on some datasets, the trend can be completely reversed on others.
This calls for more comprehensive evaluations on multiple datasets and machine learning tasks for embedding quality evaluation metrics.
Overall, only coherence provides strong signal in a single direction across all the datasets and perturbation methods.

\subsection{Metric Similarity}

Since there are no clear winners in the experiments, it is important to use multiple metrics in real-world applications.
Figure~\ref{fig:imagenet-pairplot} presents pairwise correlations and kernel densities of different metrics on the training and validation sets of ImageNet.
Overall, there are two clusters of the metrics: NESum and Stable rank as one and Coherence, $\alpha$-ReQ, RankMe and condition number in another.
 \section{Conclusions}\label{sec:conclusions}

Is it possible to estimate embedding quality based on its statistical properties?
This paper demonstrates it is possible in two scenarios outside of the known one of self-supervised learning.
We introduced four new metrics based on ideas from numerical linear algebra, analysis of linear regression and high-dimensional probability.

We conducted a large-scale study on two novel domains for unsupervised embedding quality evaluation: prediction of supervised test set performance and predicting performance of much simpler single-layer graph embedding methods.
In case of supervised models, there seem to be no one-size-fits-all dominant solution, however, we identify numerically stable metrics that have strong correlation with downstream task performance.
In the shallow model case, metrics introduced in this work show favorable downstream performance correlation consistently across 9 different datasets. 
\bibliographystyle{icml2023}
\bibliography{bibliography}

\iftrue
\newpage
\appendix
\onecolumn
\section{Appendix.}

Here we present the list of models we used for experimenting on the training and validation sets of ImageNet.

\textbf{Training set models} \\
\texttt{beitv2\_base\_patch16\_224.in1k\_ft\_in22k\_in1k}\\ \texttt{coat\_tiny}\\ \texttt{convnext\_base.fb\_in22k\_ft\_in1k\_384}\\ \texttt{convnext\_femto\_ols.d1\_in1k}\\ \texttt{dla46x\_c}\\ \texttt{edgenext\_base}\\ \texttt{edgenext\_small}\\ \texttt{edgenext\_x\_small}\\ \texttt{edgenext\_xx\_small}\\ \texttt{eva\_giant\_patch14\_560.m30m\_ft\_in22k\_in1k}\\ \texttt{eva\_large\_patch14\_196.in22k\_ft\_in22k\_in1k}\\ \texttt{eva\_large\_patch14\_336.in22k\_ft\_in22k\_in1k}\\ \texttt{lcnet\_050.ra2\_in1k}\\ \texttt{lcnet\_075.ra2\_in1k}\\ \texttt{lcnet\_100.ra2\_in1k}\\ \texttt{levit\_128s}\\ \texttt{maxvit\_base\_tf\_512.in21k\_ft\_in1k}\\ \texttt{maxvit\_large\_tf\_512.in21k\_ft\_in1k}\\ \texttt{mobilenetv3\_large\_100.miil\_in21k\_ft\_in1k}\\ \texttt{mobilenetv3\_small\_075.lamb\_in1k}\\ \texttt{mobilenetv3\_small\_100.lamb\_in1k}\\ \texttt{mobilevit\_xs}\\ \texttt{mobilevit\_xxs}\\ \texttt{mobilevitv2\_100}\\ \texttt{mobilevitv2\_150\_384\_in22ft1k}\\ \texttt{regnetz\_d8}\\ \texttt{rexnet\_100}\\ \texttt{swin\_large\_patch4\_window12\_384}\\ \texttt{tf\_efficientnet\_b0.ns\_jft\_in1k}\\ \texttt{tf\_efficientnet\_b3.ns\_jft\_in1k}\\ \texttt{tf\_efficientnet\_b4.ns\_jft\_in1k}\\ \texttt{tf\_efficientnet\_b5.ns\_jft\_in1k}\\ \texttt{tf\_efficientnet\_b6.ns\_jft\_in1k}\\ \texttt{tf\_efficientnet\_b7.ns\_jft\_in1k}\\ \texttt{tf\_efficientnetv2\_b0.in1k}\\ \texttt{tf\_mobilenetv3\_small\_100.in1k}\\ \texttt{tinynet\_e.in1k}\\ \texttt{vit\_base\_patch16\_clip\_224.laion2b\_ft\_in12k\_in1k}\\ \texttt{vit\_base\_patch16\_clip\_384.laion2b\_ft\_in12k\_in1k}\\ \texttt{vit\_base\_patch32\_clip\_224.laion2b\_ft\_in12k\_in1k}\\ \texttt{vit\_base\_patch32\_clip\_384.laion2b\_ft\_in12k\_in1k}\\ \texttt{volo\_d1\_384}\\ \texttt{volo\_d2\_384}\\ \texttt{volo\_d3\_448}\\ \texttt{volo\_d4\_448}\\ \texttt{xcit\_nano\_12\_p8\_384\_dist}\\ \texttt{xcit\_small\_12\_p8\_384\_dist}\\ \texttt{xcit\_small\_24\_p8\_384\_dist}\\ \texttt{xcit\_tiny\_12\_p8\_384\_dist}\\ \texttt{xcit\_tiny\_24\_p8\_384\_dist}\\ 

\textbf{Validation set models} \\
\begin{longtable}{ p{.5\textwidth} p{.5\textwidth}}
\mbox{\tiny\texttt{adv\_inception\_v3}} & \mbox{\tiny\texttt{bat\_resnext26ts.ch\_in1k}} \\ \mbox{\tiny\texttt{beit\_base\_patch16\_224.in22k\_ft\_in22k}} & \mbox{\tiny\texttt{beit\_base\_patch16\_224.in22k\_ft\_in22k\_in1k}} \\ \mbox{\tiny\texttt{beit\_base\_patch16\_384.in22k\_ft\_in22k\_in1k}} & \mbox{\tiny\texttt{beit\_large\_patch16\_224.in22k\_ft\_in22k}} \\ \mbox{\tiny\texttt{beit\_large\_patch16\_224.in22k\_ft\_in22k\_in1k}} & \mbox{\tiny\texttt{beit\_large\_patch16\_384.in22k\_ft\_in22k\_in1k}} \\ \mbox{\tiny\texttt{beit\_large\_patch16\_512.in22k\_ft\_in22k\_in1k}} & \mbox{\tiny\texttt{beitv2\_base\_patch16\_224.in1k\_ft\_in22k}} \\ \mbox{\tiny\texttt{beitv2\_base\_patch16\_224.in1k\_ft\_in22k\_in1k}} & \mbox{\tiny\texttt{beitv2\_large\_patch16\_224.in1k\_ft\_in22k}} \\ \mbox{\tiny\texttt{beitv2\_large\_patch16\_224.in1k\_ft\_in22k\_in1k}} & \mbox{\tiny\texttt{botnet26t\_256}} \\ \mbox{\tiny\texttt{cait\_m36\_384}} & \mbox{\tiny\texttt{cait\_m48\_448}} \\
\end{longtable}
\fi \end{document}